\documentclass{article}
\usepackage{spconf,amsmath,graphicx}
\usepackage{acro}
\usepackage{xcolor}
\usepackage{amssymb}
\usepackage{subcaption}
\usepackage{booktabs}
\usepackage{tabularx}
\usepackage{hyperref}
\usepackage[capitalise]{cleveref}
\usepackage{enumitem}
\usepackage{pgfplots}
\usepackage{xcolor}
\usepackage{algorithm}
\usepackage{algpseudocode}
\usetikzlibrary{calc}
\usepackage{enumitem}
\setlist{nosep, leftmargin=14pt}
\usepackage{mwe} 
\usepackage{hyperref}
\usepackage{multicol}
\usepackage{xspace}

\definecolor{blue1}{RGB}{220,220,230} 
\definecolor{blue2}{RGB}{200,200,220} 
\definecolor{blue3}{RGB}{180,180,210} 
\definecolor{blue4}{RGB}{160,160,200} 
\definecolor{blue5}{RGB}{100,100,160} 
\definecolor{blue6}{RGB}{50,50,110}   

\title{An Interpretable X-Ray Style Transfer via Trainable Local Laplacian 
Filter}
\name{
\begin{tabular}{c c}
Dominik Eckert$^{\star}$,\;
Ludwig Ritschl$^{\star}$,\;
Christopher Syben$^{\star}$,\;
Christian Hümmer$^{\star}$,\\
Julia Wicklein$^{\star}$,
Marcel Beister$^{\star}$,\;
Steffen Kappler$^{\star}$,\;
Sebastian Stober$^{\dagger}$
\end{tabular}
}

\address{$^{\star}$ X-ray Products, Siemens Healthineers AG, Germany \\
     $^{\dagger}$ AI Lab, Otto von Guericke University Magdeburg, Germany}

\DeclareAcronym{llf}{
	short = LLF,
	long = Local Laplacian Filter,
}

\DeclareAcronym{mlp}{
	short = MLP,
	long = Multi-Layer Perceptron,
}

\DeclareAcronym{gan}{
	short = GAN,
	long = Generative Adversarial Network,
}

\DeclareAcronym{nn}{
	short = NN,
	long = Neural Network,
}

\DeclareAcronym{sgd}{
	short = SGD,
	long = Stochastic Gradient Descent,
}

\DeclareAcronym{ssim}{
	short = SSIM,
	long = Structural Similarity Index,
}

\DeclareAcronym{mssim}{
	short = MSSIM,
	long = Mean Structural Similarity Index,
	}

\DeclareAcronym{mse}{
	short = MSE,
	long = Mean Squared Error,
}

\newcommand{\orig}{R\textbar -\xspace}

\newcommand{\orignorm}{R\textbar N\xspace}

\newcommand{\mlpremap}{M\textbar -\xspace}

\newcommand{\mlpnorm}{M\textbar N\xspace}

\newcommand{\gradientremap}{$\nabla$-H\xspace}

\DeclareAcronym{mbtst}{
	short = MBTST,
	long = Malmö Breast Tomosynthesis Screening Trial,
	}

\DeclareAcronym{lut}{
	short = LUT,
	long = Look-Up Table,
	}

\DeclareAcronym{rm}{
	short = RM,
	long = Remapping Function,
	}

\DeclareAcronym{relu}{
	short = ReLU,
	long = Rectified Linear Unit,
	}

\DeclareAcronym{norm}{
	short = NormL,
	long = Normalization Layer,
	}

\begin{document}

\maketitle
\begin{abstract}
	Radiologists have preferred visual impressions or 'styles' of X-ray 
	images that are manually adjusted to their needs to support their
	diagnostic performance.
	In this work, we propose an automatic and interpretable X-ray style 
	transfer by introducing a trainable version of the \ac{llf} 
	\cite{paris2011local}.
	From the shape of the \ac{llf}'s optimized remap function,
	the characteristics of the style
	transfer can be inferred and reliability of the algorithm can be 
	ensured.
	Moreover, we enable the \ac{llf} to capture complex X-ray style features 
	by replacing the remap function with a \ac{mlp} and adding a trainable 
	normalization layer.
	We demonstrate the effectiveness of the proposed method by transforming 
	unprocessed mammographic X-ray images into images that match the style
	of target mammograms and achieve a \ac{ssim} of 0.94 compared to 0.82 of 
	the baseline \ac{llf} style transfer method from \cite{aubry2014fast}.

\end{abstract}

\newcommand{\remapb}{\textrm{r}(\cdot)}
\newcommand{\gauss}{\mathcal{G}}
\newcommand{\gp}{\mathcal{G}(\textrm{p})}
\newcommand{\img}{\mathbf{I}}
\newcommand{\remap}{\textrm{r}}
\newcommand{\il}{\mathcal{L}'}
\newcommand{\pil}{\mathcal{L}'(\textrm{p})}
\newcommand{\Lap}{\mathcal{L}}
\newcommand{\Lapg}{\mathcal{L}(\textrm{p})}
\newcommand{\model}{\mathbf{M}}
\newcommand{\pix}{\textrm{i}}
\newcommand{\pixd}{\delta_{p}}
\newcommand{\mlp}{\textrm{m}(\cdot)}
\newcommand{\rd}{\textrm{sign}\left(\pixd \right)  \cdot \sigma  \cdot \left( 
\frac{|\pixd |}{\sigma} \right)^\alpha
}
\newcommand{\re}{\textrm{sign} \left( \pixd \right) \cdot \left( \beta  \cdot ( 
| \pixd | - \sigma  ) + \sigma \right)}


\section{Introduction}
\label{sec:intro}
The human eye is unable to simultaneously perceive the entire spectrum of
signals acquired by an X-ray detector. Consequently, the signal must be
compressed in to a visible range of approximately 500-1000 shades of gray
\cite{barten1993spatiotemporal}. Inevitably, information is lost during this
compression. At the same time crucial diagnostic information is often contained
in subtle changes~\cite{bruno256}. Due to the ambiguity of this complex task,
various processing algorithms have emerged over time, each producing unique
X-ray image impressions, also referred to as 'style'. In addition, radiologists
have distinct preferences for X-ray image styles, shaped by their training,
personal inclinations, and neurophysiological processes
\cite{bruno256}. Modifications to the 
X-ray image impression can affect their diagnostic process, as 60-80\% of errors
are attributed to perceptual mistakes \cite{waite2017interpretive}.
Nonetheless, radiologists encounter variations in X-ray machines, equipment
modifications, and improvements in image processing pipelines, continuously.
This difficulty is mitigated by X-ray System Vendors, who manually adjust
pipeline parameters, to align with the preferences of radiologists based on
existing acquisitions with the preferred style.

We posit that, in the era of machine and deep learning, adapting an X-ray image
style to a target image style can be automated. Various methods have been
proposed to transfer the presentation of a medical image from one domain to
another. The most prominent approaches for domain transfer in medical imaging
are \ac{gan}-based methods
\cite{armanious2020medgan,tmenova2019cyclegan,kong2021breaking} and diffusion
models \cite{ozbey2023unsupervised,kim2024adaptive}.
However, manipulating X-ray image features can inadvertently remove diagnostic
information or introduce artifacts, especially when complex \acp{nn} are
involved, a scenario that might negatively impact clinical practice.

A more reliable approach is offered by \cite{aubry2014fast},
who proposed a photographic style transfer method based on the \ac{llf} by
\cite{paris2011local}. The \ac{llf} is a powerful image processing algorithm
used to enhance or diminish image structures according to their local contrast
and continues to be utilized in recent studies \cite{zhang2024lookup,avci2023effect}.
Its functionality
is entirely governed by a \ac{rm} $\remap(\cdot): \mathbb{R} \rightarrow
\mathbb{R}$. Consequently, the shape of $\remapb$ provides insights into the feature
alterations performed by the \ac{llf}.
\cite{aubry2014fast} suggests employing the \ac{llf} for style transfer by
determining $\remapb$ such that the \ac{llf} matches the gradient histograms of the
input and target images.

While this approach is promising, it should be noted that gradient histograms
do not fully encapsulate the style of an X-ray image. This raises the
question of how to capture and transfer the subtle details of X-ray image
styles while maintaining interpretability and reliability. Instead of relying
on handcrafted features like gradient histograms, we propose to let the feature
manipulation for style transfer be learned by the \ac{llf} itself. To achieve
this, we propose a trainable \ac{llf}, which can be optimized to transform
input images to match a specific target style. Moreover, we propose to enhance
the \ac{llf} to enable more complex and subtle style changes: We replace the
\ac{rm} $\remapb$ of the \ac{llf}, which is defined by only three
parameters, with a more complex \ac{mlp} $\mathrm{m}(\cdot): \mathbb{R}
\rightarrow \mathbb{R}$. This \ac{mlp} maintains the capability to map an input
scalar to an output scalar, thus preserving interpretability, while also
enhancing the performance of the \ac{llf}. Additionally, we attach a trainable
\ac{norm} to the output of the \ac{llf}, to compensate missing
functionality of the \ac{llf} in relation to X-ray images, resulting in the
two-layer image processing architecture depicted in \cref{fig:llf_optim}.
Finally, we demonstrate the effectiveness of our method in transforming
unprocessed mammographic X-ray images to a target style, comparing it to the
baseline \ac{llf} style transfer method by \cite{aubry2014fast}.

\section{Method}
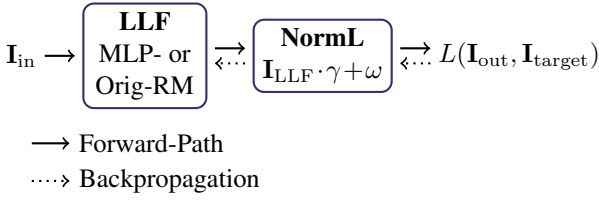
\begin{figure}
\begin{tikzpicture}[
	network/.style={rectangle, anchor=center, draw=blue6, rounded corners, 
	text centered, minimum height=2em, thick},
	textb/.style={text centered, minimum height=0.0em, inner sep=0},
	arrow/.style={->, thick},
	dottedarrow/.style={->, thick, dotted},
    ]
  
	\node[textb] (input) at (0,0) {$\mathbf{I}_\mathrm{in}$};
    \node[network, anchor=west] (llf) at ($(input.east) + (0.6, 0)$)
	{\parbox{1.4cm}{\centering \textbf{LLF}\\ \ac{mlp}- or Orig-RM}};
    \node[network, anchor=west] (norm) at ($(llf.east) + (0.6, 0)$) 
	{\parbox{1.6cm}{ \centering \textbf{\ac{norm}} \\ 
	$\mathbf{I}_\mathrm{LLF} \cdot \gamma + \omega$}};

    \node[textb, anchor=west] (l1) at ($(norm.east) + (0.6, -0.0)$) 
	{$L(\img_{\mathrm{out}}, \img_{\mathrm{target}}) $};
  
    \draw[arrow] ([xshift=0.1cm]input.east) -- ([xshift=-0.1cm]llf.west);

    \draw[arrow] ([xshift=0.1cm,yshift=1mm]llf.east) -- ([xshift=-0.1cm,yshift=1mm]norm.west);
    \draw[dottedarrow] ([xshift=-0.1cm,yshift=-1mm]norm.west) -- ([xshift=0.1cm,yshift=-1mm]llf.east);

    \draw[arrow] ([xshift=0.1cm, yshift=1mm]norm.east) -- ([xshift=-0.1cm,yshift=1mm]l1.west);
    \draw[dottedarrow] ([xshift=-0.1cm, yshift=-1mm]l1.west) -- ([xshift=0.1cm, yshift=-1mm]norm.east);

    \node[anchor=north west, align=left] at (0, -0.8) {
        \begin{tikzpicture}
            \draw[arrow] (0,0) -- (0.5,0) node[right] {Forward-Path};
            \draw[dottedarrow] (0,-0.5) -- (0.5,-0.5) node[right] {Backpropagation};
        \end{tikzpicture}
    };

    \end{tikzpicture}
	\caption{Optimization of \ac{llf} and a \acf{norm}. Either an \ac{mlp} 
	or the \ac{rm} of \cite{paris2011local} is employed.}
\label{fig:llf_optim}

\end{figure}

The objective of this study is to devise an automated style transfer for X-ray 
images. To enable an interpretable and reliable operation of the processing
algorithm in clinical practice, we propose a trainable and enhanced version of 
the \ac{llf} \cite{paris2011local} fitting for style transfer on X-ray images. 
\cref{fig:llf_optim} illustrates the complete optimization pipeline, with each 
component described in the following sections.

\subsection{\ac{llf} Algorithm}
Since the \ac{llf} is central to our approach, we first provide an overview of 
its key aspects relevant to this work. For a detailed description of the 
\ac{llf} and its complexity, refer to \cite{paris2011local} and 
\cite{aubry2014fast}.

The \ac{llf} is a non-linear image processing algorithm that utilizes Gaussian 
and Laplacian pyramid decompositions. It facilitates the manipulation of image 
features based on their local context, i.e., differences to neighboring pixels. 

In the \ac{llf}, the local context is aggregated by decomposing an input image 
$\img$ into a Gaussian pyramid $\gauss$, where the local context is represented 
by a value $\gp$ at position $\mathrm{p}$ in this pyramid.
An output pixel $ \pix' $ is then calculated from the respective input pixel
$ \pix$ by the following nonlinear function:
\begin{equation}
	\pix' = \remap(\pixd) + \gp \; ,
\end{equation}
with $\pixd = \pix - \gp$ being the difference of $\pix$ to its local context
$\gp$.
It is important to note that the \acf{rm} $\remap(\cdot): \mathbb{R} \to 
\mathbb{R}$
is the only adjustable component of the \ac{llf} and determines the
behavior of the algorithm.

\begin{figure}[tb]
	\begin{subfigure}[t]{0.49\linewidth}
	\centering
	\definecolor{color1}{RGB}{58,80,138}
\definecolor{color2}{RGB}{129,161,193}

\begin{tikzpicture}

\pgfmathsetmacro{\alpha}{0.3}
\pgfmathsetmacro{\beta}{2.0}
\pgfmathsetmacro{\sigmap}{0.2}
\pgfmathsetmacro{\ymin}{-0.5}
\pgfmathsetmacro{\ymax}{0.5}

\begin{axis}[
    xlabel={$\pixd$},
    ylabel={$\remap(\pixd)$},
    ymin=\ymin, ymax=\ymax,
    xmin=-0.5, xmax=0.5,
    xtick={-0.4, 0, 0.4},
    axis lines=left,
    width=1.0\linewidth,
    legend style={at={(0.5, 1.5)}, anchor=north, legend columns=1, font=\scriptsize},
]
\addplot [domain=-\sigmap:\sigmap, samples=100, smooth, ultra thick, color=blue2] 
	{sign(x) * \sigmap * (abs(x) / \sigmap)^\alpha };

	\addplot [domain=-0.5:-\sigmap, samples=1000, smooth, ultra thick,
	color=color1]
	{sign(x) * (\beta * (abs(x) -\sigmap) + \sigmap)  };
	\addplot [domain=\sigmap:0.5, samples=1000, smooth, ultra thick,
	color=color1]
	{sign(x) * (\beta * (abs(x) -\sigmap) + \sigmap)  };

	\addplot[dashed, color=black] coordinates {(-\sigmap, \ymin) (-\sigmap, -\sigmap)};
	\addplot[dashed, color=black] coordinates {(\sigmap, \ymin) (\sigmap, \sigmap)};
	\addplot[dashed, color=black] coordinates {(0, \ymin) (0, 0)};

\node at (axis cs:-\sigmap/2,\ymin/2) [anchor=north] {\(\sigma\)};
\draw[<->, thick] (axis cs:-\sigmap,\ymin/2) -- (axis cs:0,\ymin/2);

\node at (axis cs:\sigmap/2,\ymin/2) [anchor=north] {\(\sigma\)};
\draw[<->, thick] (axis cs:0,\ymin/2) -- (axis cs:\sigmap,\ymin/2);

\end{axis}
\end{tikzpicture}
	\caption{$ \alpha < 1$ and $ \beta > 1$}
	\label{fig:remap1}
	\end{subfigure}
	\begin{subfigure}[t]{0.49\linewidth}
	\centering
	\definecolor{color1}{RGB}{58,80,138}
\definecolor{color2}{RGB}{129,161,193}

\begin{tikzpicture}

\pgfmathsetmacro{\alpha}{2.0}
\pgfmathsetmacro{\beta}{0.5}
\pgfmathsetmacro{\sigmap}{0.2}
\pgfmathsetmacro{\ymin}{-0.5}
\pgfmathsetmacro{\ymax}{0.5}

\begin{axis}[
    ytick=\empty,
    axis y line=none,  
    xlabel={$\pixd$},
    ylabel={Y},
    ymin=\ymin, ymax=\ymax,
    xmin=-0.5, xmax=0.5,
    xtick={-0.4, 0, 0.4},
    axis lines=left,
    width=1.0\linewidth,
    legend style={at={(0.5, 1.5)}, anchor=north, legend columns=1, font=\scriptsize},
]
\addplot [domain=-\sigmap:\sigmap, samples=100, smooth, ultra thick, color=blue2] 
	{sign(x) * \sigmap * (abs(x) / \sigmap)^\alpha };

	\addplot [domain=-0.5:-\sigmap, samples=1000, smooth, ultra thick,
	color=color1]
	{sign(x) * (\beta * (abs(x) -\sigmap) + \sigmap)  };
	\addplot [domain=\sigmap:0.5, samples=1000, smooth, ultra thick,
	color=color1]
	{sign(x) * (\beta * (abs(x) -\sigmap) + \sigmap)  };

	\addplot[dashed, color=black] coordinates {(-\sigmap, \ymin) (-\sigmap, -\sigmap)};
	\addplot[dashed, color=black] coordinates {(\sigmap, \ymin) (\sigmap, \sigmap)};
	\addplot[dashed, color=black] coordinates {(0, \ymin) (0, 0)};

\node at (axis cs:-\sigmap/2,\ymin/2) [anchor=north] {\(\sigma\)};
\draw[<->, thick] (axis cs:-\sigmap,\ymin/2) -- (axis cs:0,\ymin/2);

\node at (axis cs:\sigmap/2,\ymin/2) [anchor=north] {\(\sigma\)};
\draw[<->, thick] (axis cs:0,\ymin/2) -- (axis cs:\sigmap,\ymin/2);

\end{axis}
\end{tikzpicture}
	\caption{$ \alpha > 1$ and $ \beta < 1$}
	\label{fig:remap2}
	\end{subfigure}
	\caption{This figure illustrates how $\alpha$ and $\beta$ affect the 
	remapping function: $\beta$ influences the slope of the dark blue 
	segment, while $\alpha$ affects the S-shape of the light blue segment.}
	\label{fig:remap}
\end{figure}
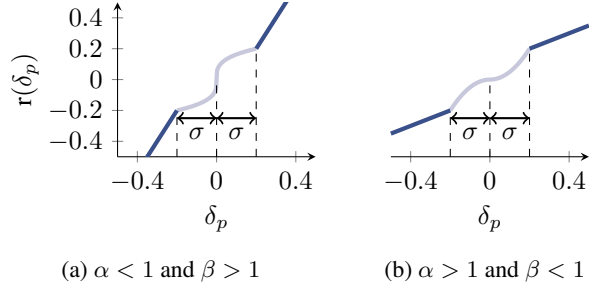

The original non-linear \ac{rm} proposed by \cite{paris2011local} comprises 
three parameters ($\sigma$, $\alpha$, and $\beta$) to adjust the behavior of the 
\ac{llf} and is defined as:
\begin{equation}
	\remap \left( \pixd \right) = \begin{cases} \rd & \text{if }  \; |\pixd| 
		< \sigma \\

	\re & \text{if } \; |\pixd| \geq \sigma \end{cases} \; ,
\end{equation}
with $\textrm{sign}(\cdot)$ being the signum and $ |\cdot| $ the
absolute value function. Different manifastation of $\remapb$ are illustrated in 
\cref{fig:remap}.

Pixels with large differences $\pixd$ from their local context can be 
interpreted as global structures, while pixels with small differences are 
considered details. To differentiate their processing, the threshold $\sigma$ 
determines whether $\pixd$ is addressed with the first term ($\alpha$) or the 
second term ($\beta$).
The parameter $\alpha$ determines the manipulation of details. If $\alpha > 1$, 
small differences are smoothed since $\pixd < \sigma$ are mapped to a smaller 
range. Conversely, if $\alpha < 1$, small differences are amplified as they are 
mapped to a larger range.
The resulting influence of $\alpha$ is depicted in light blue
in \cref{fig:remap1} and \cref{fig:remap2}.
The parameter $\beta$ influences the manipulation of global structures. If 
$\beta > 1$, large differences (i.e., $|\pixd| \geq \sigma$), such as edges, are 
amplified, as it steepens the slope of $\remapb$, as
depicted in dark blue in \cref{fig:remap}.
Conversely, if $\beta < 1$, these differences are smoothed, as the slope of 
$\remapb$ is flattened.

Thus, the behavior of the \ac{llf} is determined by the shape of the \ac{rm} and 
as illustrated in \cref{fig:remap}, this shape allows for the interpretation of 
the type of image manipulation performed by the \ac{llf}.
Furthermore, a monotonically increasing \ac{rm} preserves image information, 
adjusting the range of differences without eliminating them 
\cite{paris2011local}. Hence, verifying monotonicity after \ac{rm} optimization 
ensures the optimized \ac{llf} is reliably applicable in practical scenarios.

\subsection{Trainable \ac{llf}}
To automatically adjust the \ac{llf}, we optimize its parameters using \ac{sgd} 
and backpropagation. We implemented the \ac{llf} in PyTorch 
\cite{paszke2019pytorch}, which inherently supports automatic differentiation, 
following the algorithmic description of \cite{paris2011local}.
Additionally, we adopt the approach of \cite{aubry2014fast} by implementing a 
\ac{lut} to precalculate all possible realizations of an intermediate Laplacian 
pyramid $\Lap'$ over the range of possible $\gp$ values.
Together with training data and a loss function, this approach already allows
to optimize the parameters of the \ac{llf}.

\subsection{Enhanced Remap Function}
\label{sec:enhanced_remap}
The remap function introduced by \cite{paris2011local} is effective but limited 
in shape flexibility due to its three parameters. Given the diverse nature of 
X-ray image styles, these parameters are insufficient to allow the \ac{llf} to 
effectively match target image styles. Consequently, we propose replacing 
$\remapb$ with an \ac{mlp}, denoted as $\mathrm{m}(\cdot): \mathbb{R} \to 
\mathbb{R}$, which, like $\remapb$, operates on $ \pixd $, processing a scalar 
input to a scalar output.
On the one hand, the functionality and reliability of the \ac{llf} are 
maintained, and its interpretability is preserved by examining the shape of the 
$\mlp$ for all possible values $\pix \in [0, 1]$. On the other hand, this 
replacement enhances the flexibility of the \ac{rm}'s possible shapes and can be 
effectively optimized with backpropagation, making it a suitable component for 
the trainable \ac{llf}.
The \ac{mlp} comprises six hidden linear layers, each employing \ac{relu} 
activation \cite{nair2010rectified} and batch normalization 
\cite{ioffe2015batch}, except for the final layer. The number of neurons in the 
layers are 3, 12, 24, 24, 12, and 3, respectively.

\subsection{Normalization Layer}
\label{sec:normalization}
Unprocessed X-ray images have a wide pixel value range, requiring the X-ray 
image pipeline to compress it to a smaller range for visibility of all 
structures simultaneously. The \ac{llf}, however, is not specifically designed 
for this task, as it manipulates only relative pixel differences.
For this reason, we propose the addition of a trainable normalization layer at 
the output of the \ac{llf} to enable simple scaling and shifting of the pixel 
range. This normalization layer can be defined as:
\begin{equation}
	\img_{\textrm{norm}} = \img_{\mathrm{LLF}} \cdot \gamma + \omega \; ,
\end{equation}
where $\gamma$ and $\omega$ are the trainable parameters of the layer.
With this addition, the X-ray \ac{llf} pipeline is now complete and can 
effectively adjust the immage scale and offset based on learned parameters, as 
illustrated in \cref{fig:llf_optim}.

\subsection{Datasets}
\label{sec:datasets}

To evaluate the \ac{llf} on X-ray images, we use the open access \ac{mbtst} 
dataset, which contains mammograms from over 7325 patients. Further details on 
the dataset can be found in \cite{dahlblom_2019_mbtst-dm}. Despite this dataset 
size, we utilize only a subset of 145 images: 130 for testing and 15 for 
optimization, as the small number of parameters in the \ac{llf} requires only a 
small training set.
These images are preprocessed using a closed-source vendor pipeline to generate 
corresponding pairs of unprocessed raw projections and processed mammograms with 
a clinically relevant image impression.

\subsection{Optimization}
\label{sec:optimization}
The complete optimization is depicted in \cref{fig:llf_optim}.
The processed data from \cref{sec:datasets} is used to optimize the \ac{llf}
parameters.
All optimization processes
are conducted using the Adam optimizer \cite{kingma2014adam} with a learning
rate of 0.0001, and the \ac{llf} is trained for 300 epochs.
As a loss function, we use a combination of \ac{mse} and \ac{mssim} \cite{wang2004image}:
\begin{equation}
	\begin{split}
		L(\img_{\mathrm{out}}, \img_{\mathrm{target}}) = \, &  
		\textrm{MSE} (\img_{\mathrm{out}}, \img_{\mathrm{target}}) \, + 
		\\
	& 1 - \textrm{MSSIM}(\img_{\mathrm{out}}, \img_{\mathrm{target}})
	\end{split}
\end{equation}
We empirically found that
training with the \ac{mlp} remap function converges faster if the \ac{mlp} is
preinitialized to depict an identity function. Therefore, before optimizing on
the dataset, we optimize the \ac{mlp} to approximate the identity function.

\section{Experiments \& Results}
In this section, we evaluate our proposed method and compare it with the
gradient histogram matching baseline method from \cite{aubry2014fast}. We
conduct four experiments, designed in accordance with the configurations
outlined in \cref{tab:llf_config}. First, we demonstrate the feasibility of a
differentiable \ac{llf} by training it with its original \ac{rm} $\remapb$.
Second, we explore the benefits of substituting $\remapb$ with a \ac{mlp}.
Third, we perform an ablation study to understand the impact of a \ac{norm} on
the optimization process. The quantitative results of these experiments are
summarized in \cref{tab:llf_config}, while \cref{fig:llf_mapping} and
\cref{fig:remap_functions} showcase visual results and the optimized \acp{rm}
respectively. 

\newcommand{\llfmapp}{0.23\textwidth}
\newcommand{\llfgap}{0.5pt}
\newcommand{\llfitgap}{7.0pt}

\begin{figure*}
\begin{minipage}[b]{0.70\textwidth}
\input{figures/results_imp3.tex}
\caption{Comparison of \acs{llf}-transformed images against the target image. \\
	Additional examples are provided in the Supplementary Material on 
	\href{https://arxiv.org/abs/your_link_here}{arXiv}.}
\label{fig:llf_mapping}
\end{minipage}
\hfill
\begin{minipage}[b]{0.29\textwidth}
\centering
\begin{minipage}{0.88\textwidth}
\definecolor{color1}{RGB}{58,80,138}
\definecolor{color2}{RGB}{129,161,193}
\definecolor{color3}{RGB}{223,194,125}
\definecolor{color4}{RGB}{166,97,26}

\begin{tikzpicture}
\begin{axis}[
    xlabel={$\pixd$},
    ylabel={$\remap(\pixd)$},
    ylabel near ticks,
    yticklabel style={inner sep=0pt, outer sep=0pt},
    ymin=-2, ymax=2,
    xtick ={-0.4, 0, 0.4},
    axis lines=left,
    width=1.0\linewidth,
    legend pos=north west,
    legend style={at={(0.7, 1.0)}, anchor=north, legend columns=-1, 
	font=\scriptsize},
     legend image post style={xscale=0.5, /tikz/line width=1pt},
    ]
\addplot [smooth, ultra thick, color1] table [x=X, y=Y, col sep=comma] {images/remap/mlp.csv};
	\addlegendentry{\mlpremap}
\addplot [smooth, ultra thick, color2] table [x=X, y=Y, col sep=comma] {images/remap/mlp_offset.csv};
\addlegendentry{\mlpnorm}

\end{axis}
\end{tikzpicture}

\begin{tikzpicture}
\begin{axis}[
    xlabel={$\pixd$},
    ylabel={$\remap(\pixd)$},
    ylabel near ticks,
    yticklabel style={inner sep=0pt, outer sep=0pt},
    ymin=-2, ymax=2,
	xtick ={-0.4, 0, 0.4},
    axis lines=left,
    width=1.0\linewidth,
    legend style={at={(0.7, 0.45)}, anchor=north, legend columns=1, 
	font=\scriptsize},
     legend image post style={xscale=0.5, /tikz/line width=1pt},
]
\addplot [smooth, ultra thick, color1] table [x=X, y=Y, col sep=comma]
	{images/remap/gradient.csv};
	\addlegendentry{\gradientremap}
\addplot [smooth, ultra thick, color2] table [x=X, y=Y, col sep=comma] {images/remap/classical.csv};
	\addlegendentry{\orig \& \orignorm}

\end{axis}
\end{tikzpicture}
\end{minipage}
	\caption{Optimized \acp{rm} of the trainable \ac{llf} and the baseline 
	\gradientremap.}
\label{fig:remap_functions}
\end{minipage}
\end{figure*}

\begin{table}[tb]
\centering
\label{tab:dataset}
\setlength{\tabcolsep}{3pt}
\begin{tabularx}{\columnwidth}{l|XXXXX}
\toprule
	 &\orig & \orignorm & \mlpremap & \mlpnorm & \gradientremap \\
	\midrule
	\ac{rm} & $\remapb$ & $\remapb$ & $\mlp$ & $\mlp$ & histogram  \\
	Norm. & - & $\checkmark$ & - & $\checkmark$ & - \\
	\bottomrule
\end{tabularx}
	\caption{Overview of \ac{llf}-pipeline configurations: Each experiment 
	uses either the original \ac{rm}, \ac{mlp}, with or without a 
	normalization layer, or baseline histogram matching 
	\cite{aubry2014fast}.}
	\label{tab:llf_config}
\end{table}

\begin{table}[tb]
\centering

\begin{tabularx}{\columnwidth}{lXXXXXX}
	
\toprule
	& input & \mlpremap & \mlpnorm &
	\orig & \orignorm & \gradientremap \\
\midrule
	SSIM & 0.587 & 0.9426 & \textbf{0.9441} & 0.9190 & 0.9107 & 0.8174 \\
	MSE & 0.0270 & \textbf{0.0064} & 0.0066 & 0.0264 & 0.0105 & 0.0738 \\
\bottomrule
\end{tabularx}

\caption{Evaluation of configurations in \cref{tab:llf_config} using \ac{ssim} 
	and \ac{mse} to compare transformed images with the targets.}

\label{tab:llf_mapping_performance}
\end{table}


\cref{fig:llf_mapping} presents an image from the test set in its raw form, the
target image with the target style, and the output images of the five different
configurations of \cref{tab:llf_config}. Optimized \acp{llf} with \ac{mlp}
(\mlpnorm and \mlpremap) process the input image to closely resemble the target.
The results of the three parameter $\remapb$ and \ac{norm} (\orignorm)
closely resemble the target style. The trainable \ac{llf} with $\remapb$, but without
\ac{norm} (\orig) creates a background that is too bright and too smooth
compared to the target. Conversely, the reference method \gradientremap
overly emphasizes the dense tissues.

These observations are quantitatively supported by the results in
\cref{tab:llf_mapping_performance}. The resulting images from the test set are
compared to their corresponding target images using the \ac{ssim} and \ac{mse}.
All four configurations based on the trainable \ac{llf} outperform the baseline
method \gradientremap. Regarding \ac{ssim} performance, \mlpnorm
performs the best, followed by \mlpremap, indicating a strong influence
from the \ac{mlp}.
\mlpnorm demonstrably outperforms \orig, shown by an \ac{mse} of 0.0105
compared to 0.0264. This finding correlates with the bright background observed
in \cref{fig:llf_mapping}, implying \ac{norm}'s positive impact on the
optimization process. Despite the significant influence of scaling and offset
changes on \ac{mse}, these factors have minimal impact on the \ac{ssim}, due to
its design focus on image structural comparison rather than offset changes
\cite{wang2004image}.

Most importantly, the \acp{rm} can now be interpreted and checked for
monotonicity to ensure that no image information is lost.
\cref{fig:remap_functions} shows the optimized \acp{rm}.
All five optimized \acp{rm} demonstrated a monotonic increase, thereby ensuring
reliable application of the trained \ac{llf} models. Although non-monotonic
\acp{rm} were absent in our experiments, incorporating a regularization term
that penalizes non-monotonicity could potentially increase the likelihood of
achieving monotonic \acp{rm} in future works. Nevertheless, it is important to note that one
can always verify the monotonicity of \acp{rm} post-optimization to maintain
reliability in practical applications.
All \acp{rm}, though varied in
detail, share two major characteristics. Firstly, they possess a slope greater
than one, emphasizing large structures. Secondly, they exhibit a steep slope
around zero, suggesting an enhancement of small image details. 
The \acp{mlp} exhibit a more complex shape with a steeper slope compared  
to the $\remapb$ function, underlining the previous observation that the 
\ac{mlp} enhances the capability of the \ac{llf}.
The \acp{rm} of \orig and \orignorm have the same shape,
even though the processed results differ, indicating a strong
influence of the \ac{norm} on the optimization process. 
Lastly, the \ac{rm} of the \gradientremap is notably distinct. This
feature derives from the different objectives of the respective methods. While
the \gradientremap aims to align the gradient histograms of the images, our
proposed methodology enables the \ac{llf} to independently learn and match the
key style attributes.

\section{Conclusion}
%
%
%

Our experiments show that the trainable and enhanced \ac{llf} effectively
learns and maps raw X-ray images to specific mammographic target styles. It is
capable of capturing detailed style nuances of the target domain. Despite using
an \ac{mlp} as the mapping function, the \ac{llf} remains interpretable,
preserving information through constraints like the monotonicity
of the \ac{rm}. The differentiability of the \ac{llf} facilitates automated
style mappings and creates new opportunities for its integration as an operator
in neural networks. Moreover, future studies could explore the applicability of the
proposed method to other medical imaging techniques, such as radiography and
computed tomography, which require mapping a signal to a visual range.
Additionally, utilizing metrics like the style loss in \cite{eckert2024stylex} to
compare non-matching X-ray styles could eliminate the need for matched images.

\section{Compliance with ethical standards}
This research study was conducted retrospectively using human subject data made 
available in open access by Lund University at 
\url{https://datahub.aida.scilifelab.se/10.23698/aida/mbtst-dm}.
Ethical approval was not required as confirmed by the license attached with the 
open access data.

\section{Disclaimer}
The concepts and information presented in this paper are based
on research and are not commercially available.

\bibliographystyle{IEEEbib}
\bibliography{refs}

\end{document}


\onecolumn
\maketitle
\begin{figure}[H]
    \centering
	\includegraphics[width=0.8\textwidth]{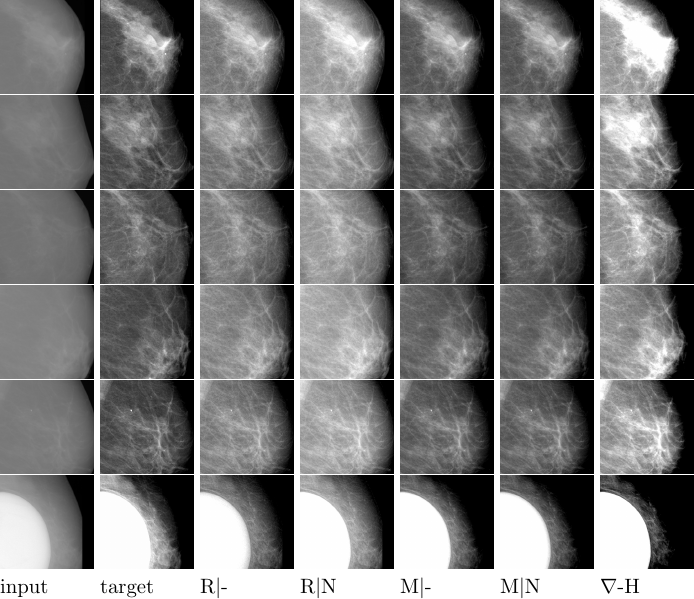}
\caption{Comparison of LLF-transformed images to the target image, including one 
	example of a mammogram with an implant.}
\end{figure}

\begin{center}
\textbf{Disclaimer:} The concepts and information presented in this paper are 
	based
on research and are not commercially available.
\end{center}